\documentclass[conference]{IEEEtran}
\IEEEoverridecommandlockouts

\usepackage{cite}
\usepackage{amsmath,amssymb,amsfonts}
\usepackage{algorithmic}
\usepackage{graphicx}
\usepackage{textcomp}
\usepackage{xcolor}
\usepackage{booktabs}
\usepackage{array}
\def\BibTeX{{\rm B\kern-.05em{\sc i\kern-.025em b}\kern-.08em
    T\kern-.1667em\lower.7ex\hbox{E}\kern-.125emX}}
\begin{document}

\title{Space Object Detection using Multi-frame Temporal Trajectory Completion Method\\
\thanks{This work is supported by the National Key Laboratory of Space Target Awareness (STA2024KGJ0202, STA2024KJW0201), the National Natural Science Foundation of China (62576281,62406249), and the Natural Science Basic Research Program of Shaanxi (2024JCYBQN-0612). Corresponding author: Han Zhang (Email: zhanghan9937@nwpu.edu.cn)}
}

\author{
\IEEEauthorblockN{Xiaoqing Lan}
\IEEEauthorblockA{\textit{School of Software} \\
\textit{Northwestern Polytechnical University}\\
Xi'an, China \\
lanxiaoqing@mail.nwpu.edu.cn}
\and
\IEEEauthorblockN{Biqiao Xin}
\IEEEauthorblockA{\textit{School of Software} \\
	\textit{Northwestern Polytechnical University}\\
	Xi'an, China \\
	biqiaoxin@mail.nwpu.edu.cn}
\and
\IEEEauthorblockN{Bingshu Wang}
\IEEEauthorblockA{\textit{School of Software} \\
\textit{Northwestern Polytechnical University}\\
Xi'an, China \\
wangbingshu@nwpu.edu.cn}
\and
\IEEEauthorblockN{Han Zhang$^\ast$}
\IEEEauthorblockA{\textit{School of Artificial Intelligence, }\\ \textit{Optics and Electronics} \\
	\textit{Northwestern Polytechnical University}\\
	Xi'an, China \\
	zhanghan9937@gmail.com} 
\and
\IEEEauthorblockN{Rui Zhu}
\IEEEauthorblockA{\textit{National Key Laboratory of}\\ \textit{Space Target Awareness} \\
	\textit{Space Engineering University}\\
	Beijing, China \\
	200347@hgd.edu.cn}
\and
\IEEEauthorblockN{Laixian Zhang}
\IEEEauthorblockA{\textit{National Key Laboratory of}\\ \textit{Space Target Awareness} \\
\textit{Space Engineering University}\\
Beijing, China \\
zhanglaixian@pku.edu.cn}
}

\maketitle
\begin{abstract}
Space objects in Geostationary Earth Orbit (GEO) present significant detection challenges in optical imaging due to weak signals, complex stellar backgrounds, and environmental interference. In this paper, we enhance high-frequency features of GEO targets while suppressing background noise at the single-frame level through wavelet transform. Building on this, we propose a multi-frame temporal trajectory completion scheme centered on the Hungarian algorithm for globally optimal cross-frame matching. To effectively mitigate missing and false detections, a series of key steps including temporal matching and interpolation completion, temporal-consistency-based noise filtering, and progressive trajectory refinement are designed  in the post-processing pipeline. Experimental results on the public SpotGEO dataset demonstrate the effectiveness of the proposed method, achieving an $F_1$ score of 90.14\%.
\end{abstract}


\section{Introduction}
\begin{figure*}[t]  
  \centering
  \includegraphics[width=\textwidth]{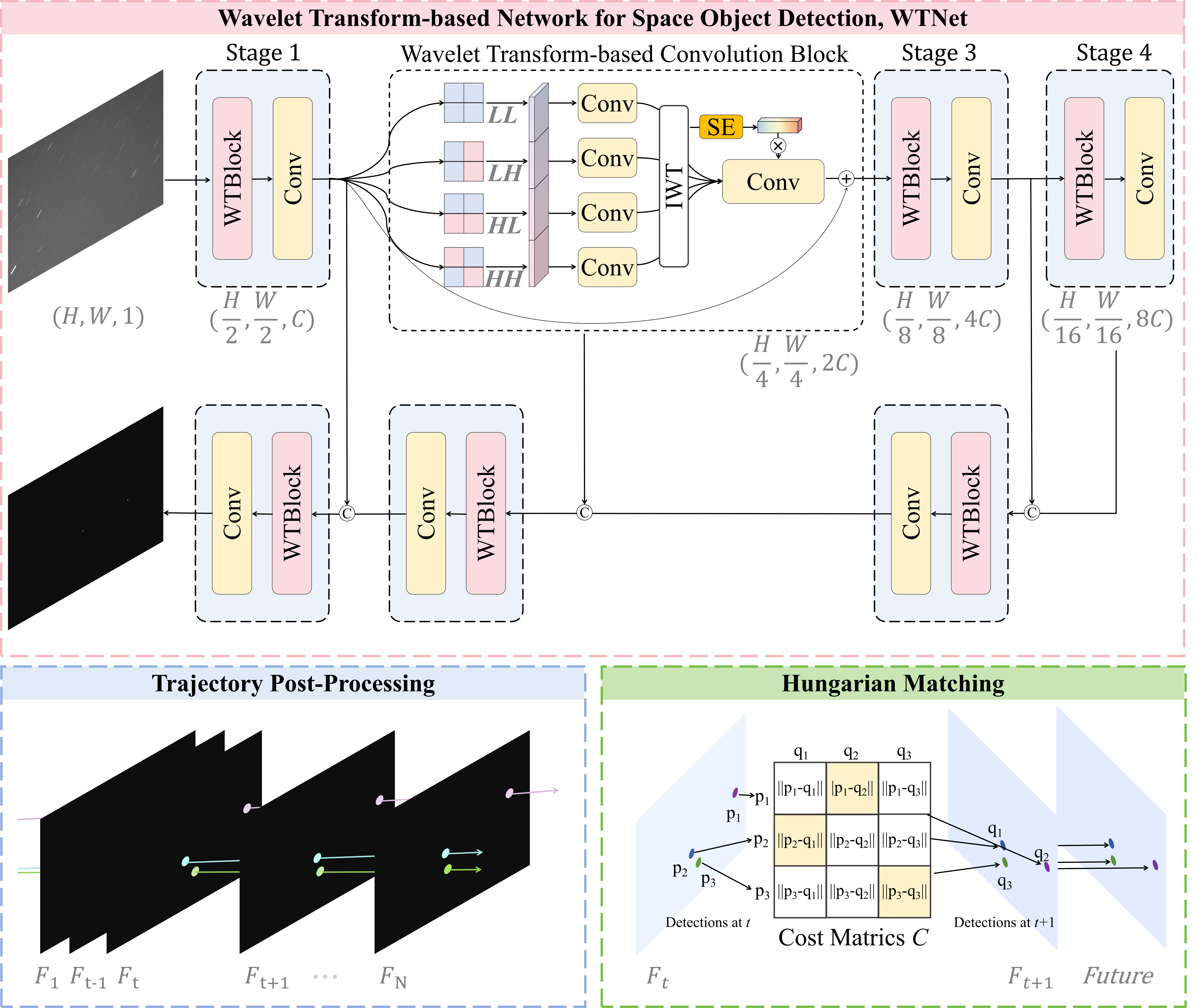}  
  \caption{The framework of the proposed method. The WTNet module is used to perform initial single-frame target detection. The multi-frame temporal trajectory completion method centered on the Hungarian algorithm is applied to filter and supplement the initial detection results.}  
  \label{fig:fig1}  
\end{figure*}
\unskip

The Geostationary Earth Orbit (GEO) constitutes a unique and crucial spatial orbital resource ~\cite{Toth2016,Abdu2021_FlexResourceOpt,Li2025_GEOSARReview,Li2025_IntellTaskSched}. However, space objects in this orbit, such as defunct satellites and discarded rocket debris, pose potential threats to space security~\cite{Fitzmaurice2021,Xue2025_SpaceObject}. Their specific imaging characteristics manifest as space objects with limited pixel occupancy, lacking discernible features in size, texture, and contour~\cite{Liu2020_TopoSweepMultiTarget}.

In 2020, the European Space Agency (ESA) and the University of Adelaide jointly launched a public challenge named SpotGEO, inviting global experts to develop computer vision algorithms for detecting space objects in low-cost telescope images~\cite{Chen2021}. The GEO object detection task is challenging due to factors such as faint target signals, stellar occlusion, and environmental interference~\cite{Wang2024_StarSuppression}.

Traditional approaches rely on image processing like transformation and energy accumulation to suppress stellar interference~\cite{Guo2021_GEO}, while recent methods adopt deep learning\cite{Wu2025_AMNet,Wang2025_LRFNet,Mao2024_SpirDet}. Representative methods include GEO-FPN with EfficientNet backbone~\cite{Abay2021_GEOFPN}, YOLO-based  object detection~\cite{Jiang2023_FaintObjects}, and hybrid designs combining single-frame detection with multi-frame post-processing~\cite{Dai2022_Multiframe}. Temporal attention mechanisms~\cite{Liu2025_TemporalAttention} and ConvLSTM frameworks~\cite{Chen2024_ConvLSTM} have further advanced multi-frame detection and tracking. Recently, attention-guided multi-task networks and end-to-end streak-like target detectors~\cite{Han2025_FaintStreaks,Zhuang2025_SDTNet} were introduced. However, most methods depend on complex models, resulting in a challenge for single-frame detection under low signal-to-noise ratio. 

Our motivation is to leverage temporal consistency priors of targets to detect space objects. It is implemented by two stages, single-frame detection and  cross-frame trajectory completion. The contributions of this paper are summarized as follows:

\begin{itemize}
\item We propose a matching-centric sequential post-processing framework. The Hungarian algorithm stabilizes cross-frame association is combined with statistical gating, neighborhood support filtering, and hierarchical completion to balance precision and recall in GEO scenarios.

\item We present an adaptive threshold estimation and gating mechanism, which integrates prior knowledge and online statistics. It enhances the method's robustness against distribution changes and device noise. 

\item Experiments on the SpotGEO dataset demonstrate that our method achieves an $F_1$ score of 90.14\%, outperforming the baseline approaches. Ablation experiments validate the fundamental role of Hungarian matching in the entire pipeline.
\end{itemize}

\section{METHODOLOGY}
The overall framework of our method is illustrated in Fig.~\ref{fig:fig1}. It first performs label transformation on the dataset, then enhances GEO target features via the WTNet module based on wavelet transform. Finally, it optimizes the single-frame detection results with the multi-frame temporal trajectory completion strategy, thereby effectively improving the performance of space object detection.

\subsection{Transformation of Dataset Labels}
The SpotGEO dataset released by the ESA provides only centroid coordinates of space objects. However, factors such as long exposure and atmospheric refraction cause targets to appear as diffuse regions. To enhance detection performance, we employ region-based fine-grained filtering for label transformation. Taking the centroid point $(x, y)$ as the reference, we extend $m$ pixels outward to construct a square local window with a side length of $2m+1$. The grayscale values within the window are normalized to the range $[0,1]$ and binarized using a threshold of 0.5. Subsequently, a circular structuring element with a radius of 1 is applied to dilate the binary mask. The resulting binary map serves as the shape label. This label covers distinguishable bright regions around the centroid while preventing excessive extension into the background area.

\subsection{Single Frame Detection Based on Wavelet-Transform Algorithm}
This paper  proposes a Wavelet Transform-based network for space object detection (WTNet) that incorporates wavelet feature enhancement during the single-frame detection stage~\cite{Xin2025_FBINet,Wang2024_YOLOv8BiFPN}. The core implementation of the network relies on the multi-scale feature processing mechanism of the wavelet feature enhancement module. First, the input features are decomposed via wavelet transform into low-frequency (LL) and high-frequency (LH, HL, HH) components. The high-frequency components undergo convolutional enhancement and scale adjustment to accentuate target edge information, while the low-frequency components preserve object contours. Subsequently, the enhanced features are reconstructed through inverse wavelet transform and fused with base convolutional features via residual connections. Finally, the CBAM attention mechanism is introduced to recalibrate the fused features along channel and spatial dimensions. The network adopts an encoder-decoder architecture overall, achieving mapping from raw images to target detection heatmaps through multi-scale feature fusion and skip connections, thereby providing more robust feature representation for space objects.

\subsection{Trajectory Processing Based on Multi-frame Temporal Completion Algorithm}

For detection tasks characterized by faint features and complex backgrounds, single-model approaches often struggle to achieve a satisfactory balance between recall and precision. Therefore, based on single-frame image detection results, we perform sequential trajectory association of candidate objects, leveraging multi-frame temporal priors to further distinguish real targets from various false detections and random noise.

First, we perform interpolation-based completion using temporal matching. For consecutive frame pairs ($f_1$,$f_2$) with detection results in the sequence, the trigger condition is set to an inter-frame interval less than 3. After obtaining the set of optimal matching pairs that minimize the total cost via the Hungarian algorithm, linear interpolation is applied to these pairs to estimate target positions within the missing intermediate frames. Defining the interpolation coefficient as $\alpha$, the interpolated position in the missing frame $f$ is:
\begin{equation}
\hat{p}_f = p_i + \alpha \cdot (q_j - p_i)
\end{equation}

To remove residual noise in the sequence, we introduce the temporal support degree principle. For a detection point $p_f$ in frame $f$ of the sequence, a temporal window is defined, and the distances between $p_f$ and detection points in each valid frame within this window are calculated. Valid support frames for $p_f$ are determined based on an inter-frame detection point distance threshold, and the support ratio is computed. If the support ratio of a detection point exceeds the specified threshold or the sequence length is relatively short, the point is retained; otherwise, it is identified as an isolated noise point and removed from the sequence.

Since prolonged target occlusion and sudden drops in signal-to-noise ratio still cause missed detections in the sequence, we adopt a progressive trajectory completion method to gradually fill in missing points from conservative to aggressive strategies. Taking the target frame as the reference, we select the preceding three frames as reference frames. Subsequently, valid matching pairs are filtered based on Euclidean distance, followed by the calculation and normalization of displacement vectors for each frame. The calculation formula is given as follows:
\begin{equation}
\vec{v} = \left( \frac{x_{\text{ref}} - x_{\text{check}}}{i}, \frac{y_{\text{ref}} - y_{\text{check}}}{i} \right)
\end{equation}

For dual-end constraint scenarios, we employ an interpolation-based completion strategy. The Hungarian algorithm is used to achieve optimal matching of detection points between the preceding and subsequent frames.The interpolated coordinates of targets in missing frames are then calculated based on temporal weighting ratios:
\begin{equation} \label{eq:interpolation}
\begin{split}
(x_{\text{interp}}, y_{\text{interp}}) &= \left( x_{\text{prev}} + \alpha(x_{\text{next}} - x_{\text{prev}}), \right. \\
&\quad \left. y_{\text{prev}} + \alpha(y_{\text{next}} - y_{\text{prev}}) \right)
\end{split}
\end{equation}

For single-end constraint scenarios, motion vector-based extrapolation is applied, where $\Delta f$ denotes the number of intervening frames, and $\vec{v}_x$, $\vec{v}_y$ represent the components of the motion vector:
\begin{equation}
(x_{\text{extrap}}, y_{\text{extrap}}) = \left( x_{\text{ref}} + \vec{v}_x \cdot \Delta f,\ y_{\text{ref}} + \vec{v}_y \cdot \Delta f \right)
\end{equation}

In case of severe data loss, linear regression is utilized to model the temporal-spatial correlation of available detections.

\begin{figure*}[t]  
  \centering
  \includegraphics[width=\textwidth]{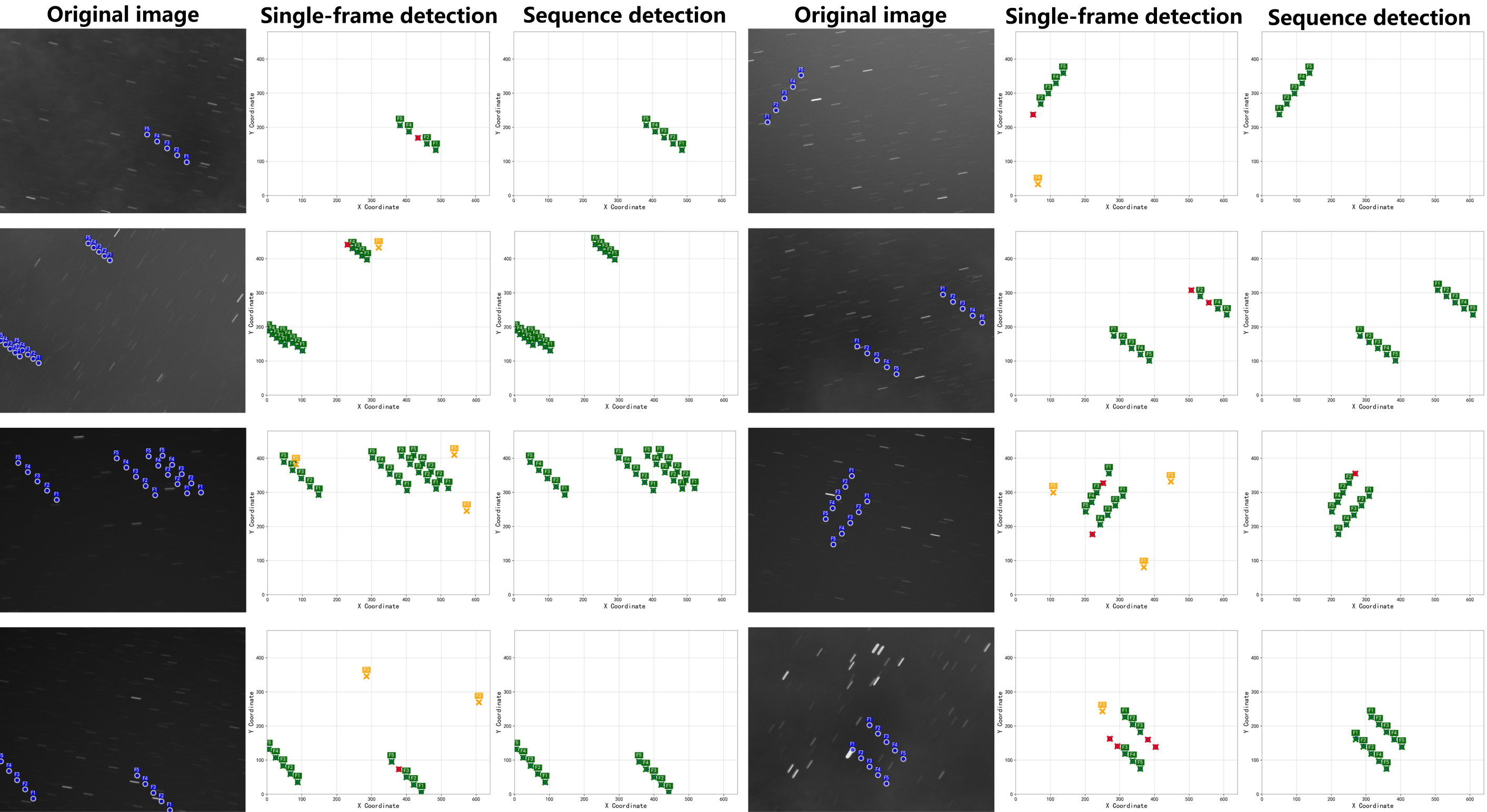}  
  \caption{ Visual results of sequential detection. The left column are original observation frames; the middle column are single-frame detection results; the right column are the final detection results after multi-frame temporal trajectory completion. The blue dots denote ground truth, the green dots represent true positives, the yellow dots indicate false positives, and red dots correspond to false negatives.}  
  \label{fig:fig2}  
\end{figure*}
\section{EXPERIMENTS}
\subsection{Experimental Setup}
Experiments were conducted on the public SpotGEO dataset. Evaluation metrics included the $F_1$ score and Mean Square Error (MSE). We implemented our model using a U-net architecture~\cite{Ronneberger2015_UNet} within the PyTorch framework. The model was trained for 400 epochs using the AdamW optimizer, with the learning rate dynamically adjusted by a cosine annealing strategy and initialized at 0.0015. All experiments were performed on an NVIDIA GeForce RTX 3090 GPU.

\subsection{Quantitative Analysis}
We conducted systematic experimental evaluations on the SpotGEO dataset and compared the results with various single-frame detection algorithms. Table I presents the performance comparison of different object detectors. It indicates that the proposed WTNet achieved an $F_1$ score of 88.07\%, while its lightweight version (WTNet-S) still attained an $F_1$ score of 85.33\%, demonstrating excellent performance stability.
\begin{table}[h]
\centering
\caption{Performance comparison of different methods on the SpotGEO dataset.}
\begin{tabular*}{\linewidth}{@{\extracolsep{\fill}}lccc}
\toprule
\textbf{Methods} & \textbf{Params(M)} & \textbf{F1(\%)} & \textbf{MSE} \\
\midrule
ResNet18       & 11.7   & 75.43 & 259633.7200 \\
Faster R-CNN   & 42  & 80.19 & 162168.6600 \\
Cascade R-CNN  & 47  & 82.54  & 177082.0900 \\
YOLOv3         & 35  & 81.89 & 141504.5600 \\
PP-YOLO       & 45     & 82.21 & 181461.1300 \\
PP-YOLOv2      & 54     & 84.08 & 160997.9800 \\
PP-YOLO-SOD  & -- & 83.97 & 140923.2000\\ 
WTNet(Ours)    & 2.7   & 88.07 & 69526.5382 \\
WTNet-S(Ours)    & 0.38   & 85.33 & 81420.6230 \\
\bottomrule
\end{tabular*}
\label{tab:addlabel}
\end{table}

To validate the effectiveness of the sequential trajectory completion module, we tested different post-processing strategies.  Experimental results indicate that combining single-frame detection with spatial point clustering alleviated false alarms to some extent, achieving an $F_1$ score of 88.56\%. However, due to the lack of a trajectory-level constraint mechanism, inter-frame mismatch remained noticeable. To address this, the Hungarian matching-based trajectory completion strategy established one-to-one correspondences between consecutive frames and performed completion for missing trajectories. This strategy ultimately enabled our method to improve the $F_1$  to 90.14\% while reducing the MSE to 61958.1973.
\begin{table}[h]
\centering
\caption{Performance comparison of different methods on the BUAA-MSOD dataset.}
\begin{tabular*}{\linewidth}{@{\extracolsep{\fill}}lcccc}
\toprule
\textbf{Methods} & \textbf{Params(M)} & \textbf{F1(\%)} & \textbf{Precision(\%)} & \textbf{Recall(\%)} \\
\midrule
DNANet          & 4.7   & 87.34 & 86.91 & 87.89 \\
HCFNet          & 15.29 & 83.33 & 74.50 & 94.65 \\
YOLOv11         & --    & 78.34 & 82.76 & 74.37 \\
WTNet(Ours)     & 2.7   & 90.88 & 87.89 & 94.08 \\
WTNet-S(Ours) & 0.38  & 89.54 & 88.67 & 90.42 \\
\bottomrule
\end{tabular*}
\label{tab:addlabel2}
\end{table}

We conducted extended experiments on the real-world ground-based astronomical observation dataset BUAA-MSOD. This dataset effectively supports space object detection studies under multi-target, low signal-to-noise ratio, and complex stellar background conditions. After retraining and testing, as shown in Table II, the proposed method achieved an $F_1$ score of 90.88\%, a precision of 87.89\%, and a recall of 94.08\% in the single-frame object detection task,outperforming several detection methods~\cite{Li2023_DenseNestedAttention,Xu2024_HCFNet,Khanam2024_YOLOv11}.  Additionally, we evaluated the performance of the lightweight model. Experimental results indicate that the model maintains strong detection performance with significant reduction in parameter count.

\subsection{Visual Results}
Fig.~\ref{fig:fig2} presents the sequential detection results of our method on the SpotGEO dataset. The selected eight sequences include both single-target and multi-target detection scenarios. Close observation reveals that during the single-frame detection stage, the wavelet transform module effectively enhances the high-frequency features of space objects, maintaining distinct separability against strong stellar backgrounds and high-noise interference, while achieving precise candidate point localization. After incorporating Hungarian matching for multi-frame trajectory fitting and completion, the system successfully restores trajectory continuity and integrity even in challenging conditions including dense star fields, low signal-to-noise ratios, and partial occlusion, significantly reducing the probabilities of missing detections.

We applied the proposed method to the BUAA-MSOD dataset and conducted a visual analysis of the detection results. As shown in Fig. 3, the first column displays the original images, the second column corresponds to the ground truth (GT) data, and the last two columns present partial sequential detection results before and after the lightweighting of the model, respectively. The results demonstrate that the proposed method maintains strong detection performance in long-sequence multi-target scenarios, further validating its generalization capability.
\begin{figure}[h]
\centering
\includegraphics[width=\linewidth]{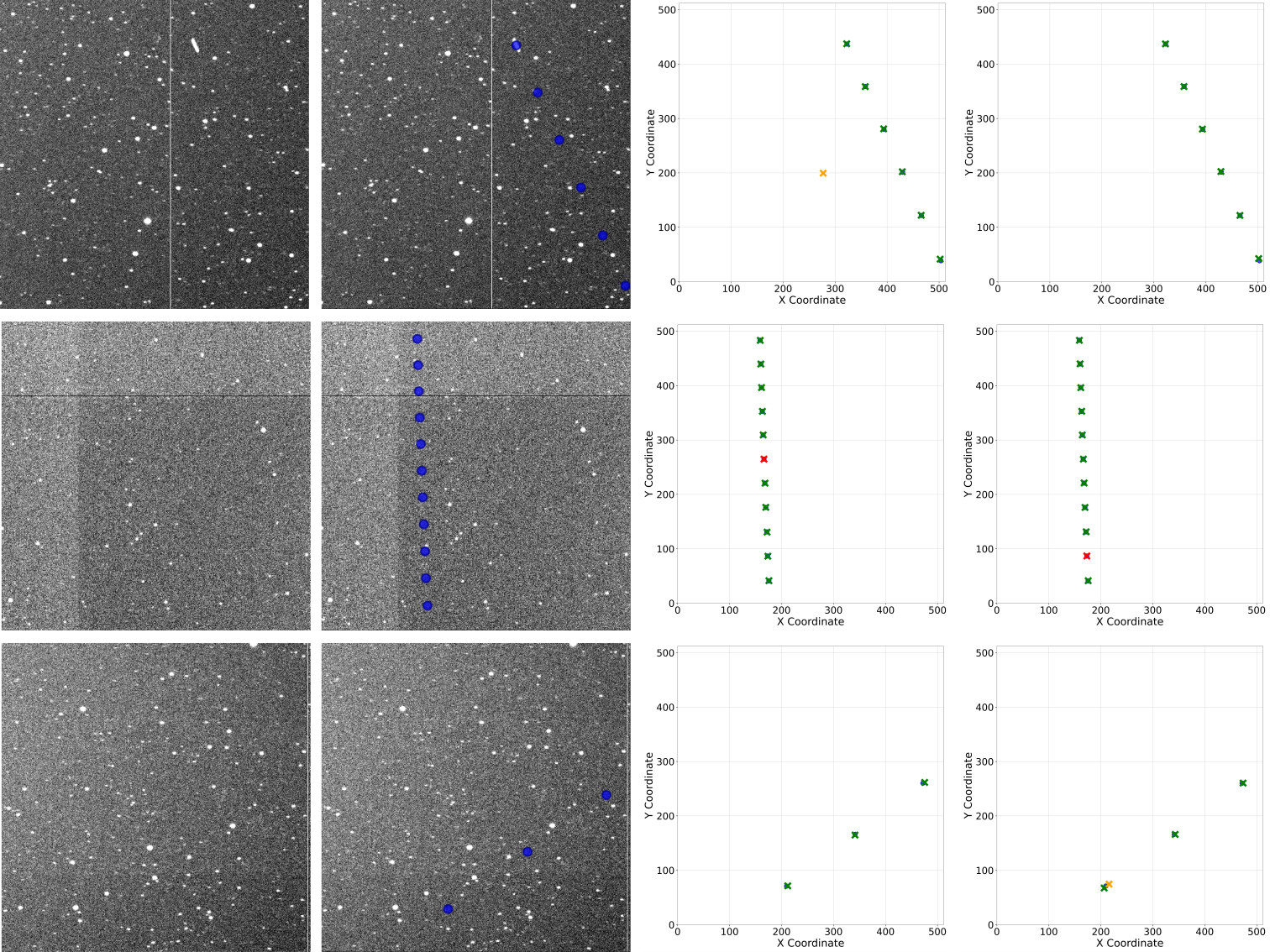}
\caption{Single-frame detection results of the proposed method on the BUAA-MSOD dataset. The first column shows the original images; the second column presents the ground truth (GT) labels; the third and fourth columns display the detection results before and after model lightweighting, respectively. Blue, green, yellow, and red points represent real targets, true positives, false positives, and false negatives, respectively.}
\label{fig:fig3}
\end{figure}

\section{CONCLUSION}
In this paper, we propose a framework based on single-frame detection and multi-frame temporal trajectory completion. The proposed WTNet achieves an $F_1$ of 88.07\%, outperforming the typical single-frame detection approaches. With the introduction of the multi-frame temporal trajectory completion strategy, the $F_1$ is improved to 90.14\%, validating the effectiveness of this approach. Extended experiments on the public BUAA-MSOD dataset further confirm the generalization capability of our method.This combination of single-frame detection and temporal completion strategy is promising to improve space object detection. In future, this can be invested to improve the model's robustness across diverse observational scenarios and complex space environments.
\bibliographystyle{IEEEtran}
\bibliography{references}

@article{Toth2016,
  author  = {Toth, C. and Józków, G.},
  title   = {Remote sensing platforms and sensors: a survey},
  journal = {ISPRS Journal of Photogrammetry and Remote Sensing},
  volume  = {115},
  pages   = {22--36},
  year    = {2016}
}

@article{Li2025_GEOSARReview,
  author    = {Li, J. and others},
  title     = {A review of recent development of geosynchronous synthetic aperture radar technique},
  journal   = {Remote Sensing},
  volume    = {17},
  year      = {2025},
  number    = {20},
  pages     = {3405},
  doi       = {10.3390/rs17203405}
}

@article{Abdu2021_FlexResourceOpt,
  author    = {Abdu, T. S. and Kisseleff, S. and Lagunas, E. and Chatzinotas, S.},
  title     = {Flexible resource optimization for GEO multibeam satellite communication system},
  journal   = {IEEE Transactions on Wireless Communications},
  year      = {2021},
  volume    = {20},
  number    = {12},
  pages     = {7888--7902},
  doi       = {10.1109/TWC.2021.3088609}
}

@article{Li2025_IntellTaskSched,
  author    = {Li, D. and Wu, S. and Wang, Y. and Wu, W. and Zhang, Q.},
  title     = {Intelligent task scheduling in hybrid GEO-LEO satellite-assisted marine IoT network},
  journal   = {IEEE Internet of Things Journal},
  year      = {2025},
  volume    = {12},
  number    = {7},
  pages     = {8353--8367},
  doi       = {10.1109/JIOT.2024.3502791}
}

@article{Fitzmaurice2021,
  author  = {Fitzmaurice, J. and Bédard, D. and Lee, C. H. and Seitzer, P.},
  title   = {Detection and correlation of geosynchronous objects in NASA’s wide-field infrared survey explorer images},
  journal = {Acta Astronautica},
  volume  = {183},
  pages   = {176--198},
  year    = {2021}
}

@article{Xue2025_SpaceObject,
  author    = {Xue, J. and Zhang, Y. and Tao, X. and Zhao, S.},
  title     = {Research on space object origin tracing approach using density peak clustering and distance feature optimization},
  journal   = {Applied Sciences},
  volume    = {15},
  year      = {2025},
  number    = {20},
  pages     = {10943},
  doi       = {10.3390/app152010943}
}

@article{Liu2020_TopoSweepMultiTarget,
  author    = {Liu, D. and Chen, B. and Chin, T.-J. and Rutten, M. G.},
  title     = {Topological sweep for multi-target detection of geostationary space objects},
  journal   = {IEEE Transactions on Signal Processing},
  year      = {2020},
  volume    = {68},
  number    = {},
  pages     = {5166--5177},
  doi       = {10.1109/TSP.2020.3021232}
}

@article{Wang2024_StarSuppression,
  author    = {Wang, X. and others},
  title     = {Star suppression based on structure of star groups for geosynchronous object detection using wide-field telescopes},
  journal   = {IEEE Transactions on Aerospace and Electronic Systems},
  year      = {2024},
  volume    = {60},
  number    = {3},
  pages     = {3160--3176},
  doi       = {10.1109/TAES.2024.3360957}
}

@inproceedings{Chen2021,
  author    = {Chen, B. and others},
  title     = {Spot the GEO satellites: from dataset to Kelvins SpotGEO challenge},
  booktitle = {Proceedings of the 2021 IEEE/CVF Conference on Computer Vision and Pattern Recognition Workshops (CVPRW)},
  address   = {Nashville, TN, USA},
  pages     = {2086--2094},
  year      = {2021},
  publisher = {IEEE}
}

@inproceedings{Guo2021_GEO,
  author    = {Guo, L. and Zhang, W. and Wang, Z. and Sun, X. and Shang, Y.},
  title     = {Weak geo satellite target detection based on image transformation and energy accumulation},
  booktitle = {Proceedings of the 4th International Conference on Image and Graphics Processing},
  year      = {2021},
  pages     = {52--58},
  publisher = {ACM},
  address   = {New York}
}

@inproceedings{Abay2021_GEOFPN,
  author    = {Abay, R. and Gupta, K.},
  title     = {GEO-FPN: a convolutional neural network for detecting GEO and near-GEO space objects from optical images},
  booktitle = {Proceedings of the 8th European Conference on Space Debris},
  year      = {2021},
  pages     = {123--129},
  address   = {Darmstadt, Germany},
  publisher = {ESA Space Debris Office}
}

@article{Jiang2023_FaintObjects,
  author    = {Jiang, Y. and Tang, Y. and Ying, C.},
  title     = {Finding a needle in a haystack: faint and small space object detection in 16-bit astronomical images using a deep learning-based approach},
  journal   = {Electronics},
  volume    = {12},
  number    = {23},
  pages     = {4820},
  year      = {2023}
}

@article{Dai2022_Multiframe,
  author    = {Dai, Y. and Zheng, T. and Xue, C. and Zhou, L.},
  title     = {Effective multi-frame optical detection algorithm for GEO space objects},
  journal   = {Applied Sciences},
  volume    = {12},
  number    = {9},
  pages     = {4610--4626},
  year      = {2022}
}

@article{Liu2025_TemporalAttention,
  author    = {Liu, J. and Yu, F. and Yuan, Y. and Yang, Y.},
  title     = {Multi-frame temporal dense nested attention method for detecting GEO objects},
  journal   = {Advances in Space Research},
  volume    = {75},
  number    = {9},
  pages     = {6911--6923},
  year      = {2025}
}

@article{Chen2024_ConvLSTM,
  author    = {Chen, S. and Wang, H. and Shen, Z. and Wang, K. and Zhang, X.},
  title     = {Convolutional long-short term memory network for space debris detection and tracking},
  journal   = {Knowledge-Based Systems},
  volume    = {304},
  pages     = {112535},
  year      = {2024}
}

@article{Han2025_FaintStreaks,
  author    = {Han, Y. and Wen, D. and Li, J. and Qiao, Z.},
  title     = {Improved detection of multiple faint streak-like space targets in a single star image},
  journal   = {Remote Sensing},
  volume    = {17},
  number    = {4},
  pages     = {631},
  year      = {2025}
}

@article{Zhuang2025_SDTNet,
  author    = {Zhuang, G. and others},
  title     = {High performance space debris tracking in complex skylight backgrounds with a large-scale dataset},
  journal   = {arXiv preprint},
  eprint    = {2506.02614},
  year      = {2025}
}

@article{Wu2025_AMNet,
  author    = {Wu, F. and others},
  title     = {Attention-guided multi-task network for streak-like dim and small space target detection in single optical images},
  journal   = {Advances in Space Research},
  year      = {2025}
}

@article{Mao2024_SpirDet,
  author  = {Mao, Q. and others},
  title   = {SpirDet: toward efficient, accurate, and lightweight infrared small-target detector},
  journal = {IEEE Transactions on Geoscience and Remote Sensing},
  volume  = {62},
  number  = {},
  pages   = {1--12},
  year    = {2024},
  doi     = {10.1109/TGRS.2024.3470514}
}

@inproceedings{Xin2025_FBINet,
  author    = {Xin, B. and Li, Q. and Mao, Q. and Wang, J. and Wang, B.},
  title     = {FBI-Net: frequency band integration network for infrared small target segmentation},
  booktitle = {ICASSP 2025 - 2025 IEEE International Conference on Acoustics, Speech and Signal Processing (ICASSP)},
  year      = {2025},
  pages     = {1--5},
  publisher = {IEEE},
  address   = {Hyderabad, India},
  doi       = {10.1109/ICASSP49660.2025.10888849}
}

@article{Wang2024_YOLOv8BiFPN,
  author    = {Wang, B. and Li, C. and Zou, W. and Zheng, Q.},
  title     = {Foreign object detection network for transmission lines from unmanned aerial vehicle images},
  journal   = {Drones},
  volume    = {8},
  number    = {8},
  pages     = {361},
  year      = {2024},
  doi       = {10.3390/drones8080361}
}

@ARTICLE{Wang2025_LRFNet,
  author={Wang, Bingshu and Xin, Biqiao and Mao, Qianchen and Zhang, Han and Zhang, Laixian and Chen, C.L. Philip and Zhao, Yue},
  journal={IEEE Journal of Selected Topics in Applied Earth Observations and Remote Sensing}, 
  title={LRFNet: Toward Accurate Infrared Small Target Segmentation via Large Effective Receptive Field With Contextual Cues}, 
  year={2025},
  volume={18},
  number={},
  pages={22013-22028},
  doi={10.1109/JSTARS.2025.3599566}
  }

@inproceedings{Ronneberger2015_UNet,
  author    = {Ronneberger, O. and Fischer, P. and Brox, T.},
  title     = {U-Net: convolutional networks for biomedical image segmentation},
  booktitle = {Medical Image Computing and Computer-Assisted Intervention -- MICCAI 2015},
  year      = {2015},
  pages     = {234--241},
  publisher = {Springer International Publishing},
  address   = {Cham}
}

@article{Li2023_DenseNestedAttention,
  author    = {Li, B. and others},
  title     = {Dense nested attention network for infrared small target detection},
  journal   = {IEEE Transactions on Image Processing},
  volume    = {32},
  number    = {},
  pages     = {1745--1758},
  year      = {2023},
  doi       = {10.1109/TIP.2022.3199107}
}

@inproceedings{Xu2024_HCFNet,
  author    = {Xu, S. and others},
  title     = {HCF-Net: hierarchical context fusion network for infrared small object detection},
  booktitle = {Proceedings of the 2024 IEEE International Conference on Multimedia and Expo (ICME)},
  year      = {2024},
  pages     = {1--6},
  address   = {Niagara Falls, ON, Canada},
  publisher = {IEEE},
  doi       = {10.1109/ICME57554.2024.10687431}
}

@article{Khanam2024_YOLOv11, 
  author    = {KHANAM, R. and HUSSAIN, M.}, 
  title     = {YOLOv11: an overview of the key architectural enhancements},
  journal   = {arXiv preprint},
  eprint    = {2410.17725},
  year      = {2024}
}
\vspace{12pt}
\color{red}
\end{document}